\documentclass[lettersize,journal]{IEEEtran}
\usepackage{amsmath,amsfonts}
\usepackage{algorithmic}
\usepackage{algorithm}
\usepackage{array}
\usepackage[caption=false,font=normalsize,labelfont=sf,textfont=sf]{subfig}
\usepackage{textcomp}
\usepackage{stfloats}
\usepackage{url}
\usepackage{verbatim}
\usepackage{graphicx}
\usepackage{cite}

\usepackage{multirow}
\usepackage{listings}
\usepackage{color}
\usepackage{xcolor}

\begin{document}

\title{M$^3$Net: Multilevel, Mixed and Multistage Attention Network for Salient Object Detection}

\author{Yao Yuan,
        Pan Gao, ~\IEEEmembership{Member,~IEEE,}
        and Xiaoyang Tan
	% <-this % stops a space
	\thanks{Y. Yuan, P. Gao, and X. Tan are with College of Computer Science and Technology, Nanjing University of Aeronautics and Astronautics, Nanjing 211106, China.}% <-this % stops a space
	}

\markboth{Journal of \LaTeX\ Class Files,~Vol.~14, No.~8, August~2021}%
{Shell \MakeLowercase{\textit{et al.}}: A Sample Article Using IEEEtran.cls for IEEE Journals}


\maketitle

\begin{abstract}
Most existing salient object detection methods mostly use U-Net or feature pyramid structure, which simply aggregates feature maps of different scales, ignoring the uniqueness and interdependence of them and their respective contributions to the final prediction. To overcome these, we propose the M$^3$Net, i.e., the Multilevel, Mixed and Multistage attention network for Salient Object Detection (SOD). Firstly, we propose Multiscale Interaction Block which innovatively introduces the cross-attention approach to achieve the interaction between multilevel features, allowing high-level features to guide low-level feature learning and thus enhancing salient regions. Secondly, considering the fact that previous Transformer based SOD methods locate salient regions only using global self-attention while inevitably overlooking the details of complex objects, we propose the Mixed Attention Block. This block combines global self-attention and window self-attention, aiming at modeling context at both global and local levels to further improve the accuracy of the prediction map. Finally, we proposed a multilevel supervision strategy to optimize the aggregated feature stage-by-stage. Experiments on six challenging datasets demonstrate that the proposed M$^3$Net surpasses recent CNN and Transformer-based SOD arts in terms of four metrics. Codes are available at https://github.com/I2-Multimedia-Lab/M3Net. 
\end{abstract}


\section{Introduction}
\IEEEPARstart{S}{alient} object detection \cite{8332961} (SOD) aims to identify the most significant objects or regions in an image and segment them. Given its widespread application in computer vision, SOD plays a critical role in various downstream tasks, such as object detection \cite{6587754}, semantic segmentation \cite{sun2020mining}, image understanding \cite{6888473}, and object discovery \cite{6630857}. Furthermore, SOD serves as a valuable reference for multimodal SOD tasks, including RGB-D SOD \cite{li2023mutual, 9789193, 9925217}, RGB-T SOD \cite{9801871, 9161021}, and Light field SOD \cite{9153018, 10168184}. 

\begin{figure}[t]
	\centering
	\begin{minipage}[t]{0.085\textwidth}
		\centering
		\includegraphics[scale=0.54]{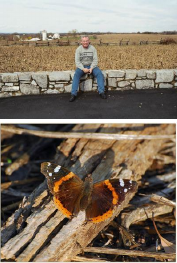}
		\centering\footnotesize{Image}
	\end{minipage}
	\begin{minipage}[t]{0.085\textwidth}
		\centering
		\includegraphics[scale=0.54]{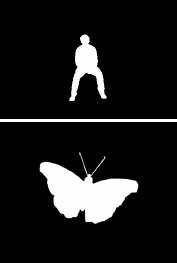}
		\centering\footnotesize{GT}
	\end{minipage}
	\begin{minipage}[t]{0.085\textwidth}
		\centering
		\includegraphics[scale=0.54]{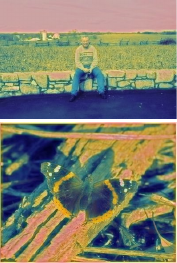}
		\centering\footnotesize{Level-1}
	\end{minipage}
	\begin{minipage}[t]{0.085\textwidth}
		\centering
		\includegraphics[scale=0.54]{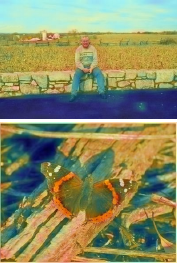}
		\centering\footnotesize{Level-2}
	\end{minipage}
	\begin{minipage}[t]{0.085\textwidth}
		\centering
		\includegraphics[scale=0.54]{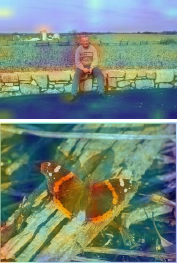}
		\centering\footnotesize{Level-3}
	\end{minipage}
	\begin{minipage}[t]{0.03\textwidth}
		\centering
		\includegraphics[scale=0.485]{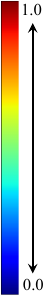}
	\end{minipage}
	\par\;\caption{The difference between multilevel features. Levels 1, 2, and 3 represent the multilevel features captured from different levels of our network. The color closer to red indicates that the model pays more attention to this area. }
	\label{fig:multilevelfeatures}
	%\vspace{-4mm}
\end{figure}

Previous CNN-based saliency detection methods have achieved significant results in positioning salient regions. Most of them take U-shape~\cite{Unet} based structures as the encoder-decoder architecture, and utilize multilevel features to reconstruct high quality feature maps in the encoder side~\cite{Amulet,8578428,8578285,Wang_2019_CVPR,BASNet,CPD}, or the decoder side~\cite{PAGRN,DSS,DHSNet,AADF,9389751,9797762,9745960}. 
For example, AADF~\cite{AADF} proposed the AD-ASPP module that combines the local saliency cues captured by dilated convolutions at a small rate and the global saliency cues captured by dilated convolutions at a large rate. DCENet~\cite{9389751} designed a dense context exploration module to capture dense multi-scale contexts, thereby enhancing the feature's discriminability. 
Nevertheless, previously mentioned methods mostly apply the same processing to features at different scales or levels, disregarding the uniqueness and interdependence between different levels of features and their distinct contributions to the final prediction. We argue that low-level features contain more non-salient information, which may harm the final prediction. As shown in Figure \ref{fig:multilevelfeatures}, lower-level feature generally contains more non-salient regions and background noises.

%In this case, we innovatively employ interactive attention to facilitate the use of complementarity between multilevel features, thereby replacing previous feature enhancement methods. 

\begin{figure}[t]
	\centering
	\begin{minipage}[t]{0.09\textwidth}
		\centering
		\includegraphics[scale=0.42]{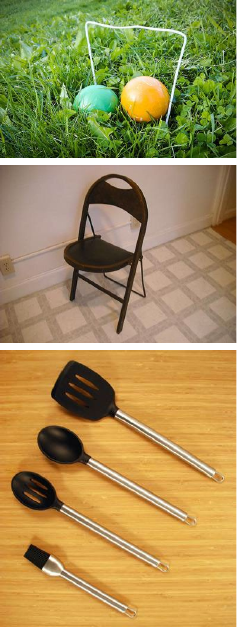}
		\centering\footnotesize{Image}
	\end{minipage}
	\begin{minipage}[t]{0.09\textwidth}
		\centering
		\includegraphics[scale=0.42]{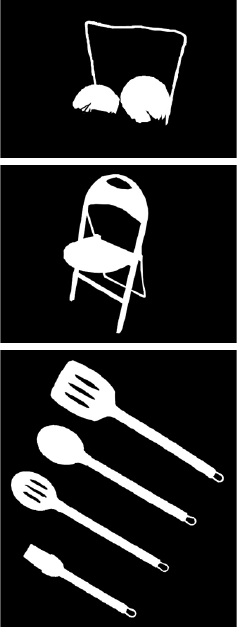}
		\centering\footnotesize{GT}
	\end{minipage}
	\begin{minipage}[t]{0.09\textwidth}
		\centering
		\includegraphics[scale=0.42]{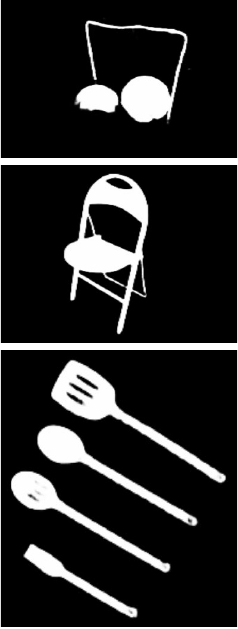}
		\centering\footnotesize{M$^3$Net}
	\end{minipage}
	\begin{minipage}[t]{0.09\textwidth}
		\centering
		\includegraphics[scale=0.42]{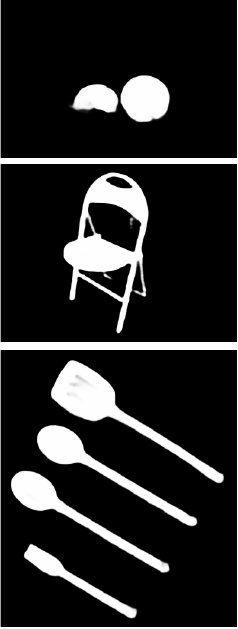}
		\centering\footnotesize{SRformer\par~\cite{SelfReformer}}
	\end{minipage}
	\begin{minipage}[t]{0.09\textwidth}
		\centering
		\includegraphics[scale=0.42]{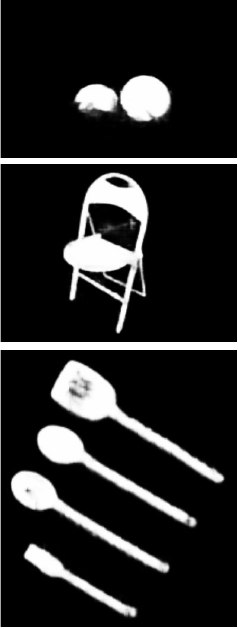}
		\centering\footnotesize{VST\par~\cite{VST}}
	\end{minipage}
	\begin{minipage}[t]{0.1\textwidth}
	\end{minipage}
	\par\;\caption{Previous Transformer-based methods often suffer from a loss of local details, whereas our proposed M3Net excels at preserving these fine-grained details. }
	\label{fig:localloss}
	%\vspace{-4mm}
\end{figure}

Recently, following the Vision Transformer’s (ViT)~\cite{Vit} success in image classification, some studies introduce the Transformer for dense prediction tasks, e.g., semantic segmentation~\cite{SETR, xie2021segformer, Zhang_2022_CVPR} or SOD~\cite{VST, SelfReformer, Tang2023RTransNet}. Thanks to the ability of the Transformer to quickly establish long-term dependencies, previously mentioned Transformer-based SOD methods excel in locating salient regions compared to CNN alternatives. However, the exclusive employment of global self-attention can result in the loss of numerous local details, as depicted in Figure \ref{fig:localloss}. In concealed object detection, UCT~\cite{10006834} employed CNNs to address the absence of detailed information. However, in the field of SOD, compensating for the loss of detailed information in transformers has not been fully investigated. 

In this paper, we propose the M$^3$Net for SOD. 
%Based on the U-shape~\cite{Unet} structure, we optimize and integrate multilevel features step by step. 
To facilitate the use of complementarity between multilevel features, we propose the Multilevel Interaction Block (MIB), which introduces the cross-attention mechanism to achieve the interaction of multilevel features, letting high-level features guide low-level features to strengthen salient regions. Inspired by the success of window self-attention~\cite{Swin}, which computes self-attention within local windows, we propose the Mixed Attention Block (MAB) that integrates global self-attention and window self-attention. On the basis of the localized salient regions, our MAB can model context at both global and local levels, thus enhancing final accuracy of the prediction. Our proposed MIB and MAB blocks work collaboratively to exploit multilevel features, which is different from prior works that only rely on multilevel feature concentration or use separate CNN networks for local detail learning.

Our main contributions can be summarized as follows:
\begin{itemize}
	\item Based on the feature pyramid structure, we rethink the uniqueness and independence between multiscale features and their distinctive contributions to final prediction, and accordingly propose a multiscale and multistage decoder. The decoder can optimize and integrate features step by step, and progressively reconstruct the saliency map.
	\item We introduce the Multilevel Interaction Block (MIB) to facilitate the interaction between features at different levels, specifically utilizing high-level features to guide the learning of low-level features. As demonstrated later, the MIB effectively enhances the salient region within the low-level features. 
	\item We propose the Mixed Attention Block (MAB), which integrates self-attention and window self-attention mechanisms. The MAB enables improved localization of salient regions while preserving fine-grained object details. 
	
\end{itemize}
Our proposed method differs distinctly from existing works in two key aspects. 
Firstly, in our multistage decoder, we adopt a sequential processing approach for features instead of the conventional two-dimensional form. As a result, no convolutional operations were incorporated into the decoder. 
%During the upsampling process, we apply the Fold with overlap upsampling instead of the commonly employed Bilinear approach used in previous methods. Moreover, our decoder does not incorporate any convolutional operations. 
%Secondly, we utilize the complementary advantages of multilevel features to enhance their overall quality. We take into account their uniqueness and correlation, which have been overlooked in previous feature enhancement methods. 
Secondly, we capture the global and fine-grained features both through the attention mechanisms instead of adjusting the kernel size as done in many existing methods. 
Besides, We evaluate our model on six challenging datasets, and the results show that the M$^3$Net achieves state-of-the-art results compared to recent competitive models. 
%Our model is also computationally efficient and can achieve an inference speed of 90 fps. 

\section{Related Work}

\subsection{Convolution Based Methods}
CNN-based methods are widely used in SOD and gain commendable performance, which usually takes the pre-trained networks~\cite{ResNet,VGG} as the encoder, and most of the efforts are then to design an effective decoder to conduct multilevel feature aggregation. Most methods~\cite{MiNet,Amulet,GateNet} use the U-shape~\cite{Unet} based structures as the encoder-decoder architecture and progressively aggregate the hierarchical features for final prediction, while DSS~\cite{DSS} follows the HED~\cite{HED} structure to fully utilize multilevel and multiscale features. Some methods tend to utilize the attention mechanism to learn more discriminative features, such as pixel-wise contextual attention~\cite{PiCANet}, spatial and channel attention~\cite{PAGRN}. Multi-task learning is also widely used for SOD, including with the tasks, eye fixation prediction~\cite{ASNet,Saliency_unified}, edge detection~\cite{BASNet,EGNet,wei2020label,9036909}, and image caption~\cite{Capsal}. 

Unlike previous CNN-based methods, our M$^3$Net can handle features extracted by Transformer as the encoder and performs sequential processing of feature maps in the decoder. We adopt the U-shape structure and consider the uniqueness and interdependence of multilevel features by implementing the proposed Multilevel Interaction Block, thus facilitating the subsequent aggregation of multilevel features. In addition, we employ the "fold with overlap" as the upsampling method for our token-based model and perform a comparison with other existing upsampling methods. 

\begin{figure*}[t]
	\centering % 图片居中
	\includegraphics[scale=0.65]{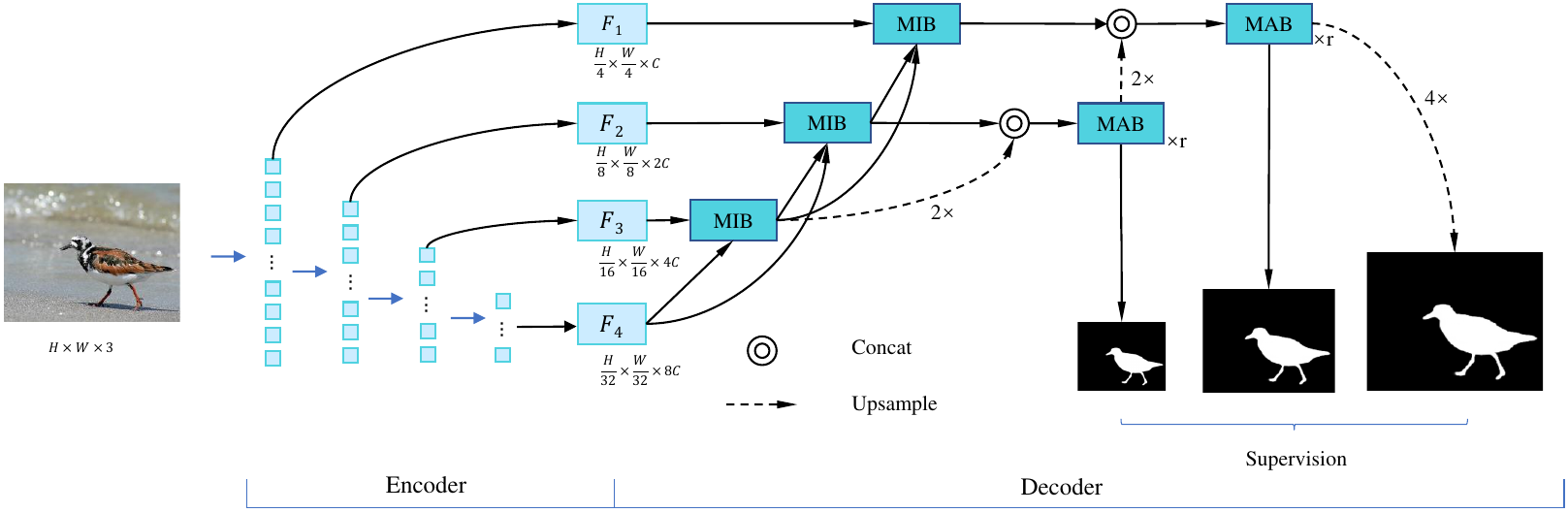}
	\caption{Overall architecture of our proposed M$^3$Net model for salient object detection. 
		The backbone is defined as a hierarchical network structure (e.g. ResNet\cite{ResNet}, SwinTransformer\cite{Swin}). 
		%The multilevel features extracted by the backbone are denoted as F$_0$ - F$_4$ and the spatial dimensions decrease sequentially. 
	}
	\label{fig:model}
	%	\vspace{-4mm}
\end{figure*}
\subsection{Transformer Based Methods}
Transformer encoder-decoder architecture was first proposed by Vaswani \textit{et al.}~\cite{Attentionisallyouneed} for natural language processing. Since the success of ViT~\cite{Vit} in image classification, more and more works have introduced Transformer architecture to computer vision tasks. SETR~\cite{SETR} and PVT~\cite{wang2021pyramid} use ViT as the encoder for semantic segmentation. 
In SOD, VST~\cite{VST} employs T2T-ViT~\cite{Yuan_2021_ICCV} as the backbone and proposes an effective multi-task decoder for features in a sequence form. 
SRformer~\cite{SelfReformer} adopted PVT as the encoder backbone and uses pixel shuffle as the upsampling method. 
%PDRNet\cite{9745960} designed a progressive dual-attention residual network to exploit two complementary attention maps, thus progressively refining prediction in a coarse-to-fine manner. 
Besides, HRTransNet~\cite{Tang2023RTransNet} explored the application of transformer in two-modality SOD tasks, such as RGB-D SOD and RGB-T SOD. 

Previous Transformer-based methods are superior in locating and capturing the salient areas in the images, while the details at the local level could be ignored. Inspired by the success of window self-attention~\cite{Swin}, we introduce window self-attention in our decoder to strengthen the ability to model local context. As shown in Figure \ref{fig:localloss}, our M$^3$Net can model context at both global and local levels, thus effectively preserving local details on the basis of localized salient objects. 
\subsection{Multilevel Feature Enhancement and Aggregation}
Multilevel feature enhancement and aggregation are crucial for gaining a high-resolution saliency map, and numerous methods have thoroughly investigated them. 
GateNet\cite{GateNet} adopted the gate mechanism to balance the contribution of each encoder block and reduce non-salient information. 
MiNet~\cite{MiNet} proposed an interactive learning method for multilevel features, aiming to minimize the discrepancies and improve the spatial coherence of multilevel features. 
ICON\cite{ICON} incorporated convolution kernels with different shapes to enhance the diversity of multilevel features. 
BBRF\cite{BBRF} designed a switch-path decoder with various receptive fields to deal with large or small-scale objects. 

To fully utilize multi-scale features, 
we first design the proposed Multiscale Interaction Block which introduces the cross-attention mechanism to achieve the interaction among multilevel features. 
In this mechanism, high-level features guide low-level feature learning from top to bottom. 
Then, in order to effectively integrate salient information across various scales within the aggregated features, 
we propose the Mixed Attention Block, which combines global self-attention and local window self-attention. 
This combination is aimed at modeling context at both global and local levels for further improvement in the accuracy of the prediction map. 

\section{Proposed method}
\subsection{Overview}
The overall architecture of our proposed M$^3$Net is given in Figure \ref{fig:model}, which includes a Transformer encoder plus a multistage decoder with mixed attention. The encoder uses the Swin Transformer as the backbone for the demonstration of the proposed methodology, while any other hierarchical network models are applicable to extract multilevel features. The decoder optimizes and integrates multilevel features step by step, and gradually reconstructs the saliency map.

After obtaining multilevel features by the backbone, we optimize the features by passing them through the multiscale interaction block, which can enhance salient regions in low-level features. Then, on the basis of detected salient regions, we further use the mixed attention block to improve the local-level fine details of the saliency map. In addition, supervision is applied to each decoder level, aiming to facilitate the model training and improve the performance of the model. 
\subsection{Swin Transformer Encoder}
Transformer based on the standard architecture~\cite{Vit} has made competitive achievements in image classification, where the relationships between a token and all other tokens are computed. However, the vanilla Transformer has quadratic complexity. This limits it to be applied to many other vision problems such as SOD, which requires tremendous tokens for dense prediction or gaining a high-resolution image. 

For efficient modeling, Swin Transformer uses window self-attention to replace standard global self-attention, reducing complexity to linear. To achieve information exchange among non-overlapping windows, shifted window partitioning is adopted, and the successive Swin Transformer blocks can be formulated as:
\begin{equation}
	\small
	\begin{split}
		&	\hat{\text{z}}^l = \text{W-MSA}(\text{LN}(\text{z}^{l-1}))+\text{z}^{l-1},\\
		&	\text{z}^l = \text{MLP}(\text{LN}(\hat{\text{z}}^l))+\hat{\text{z}}^l,\\
		&	\hat{\text{z}}^{l+1} = \text{SW-MSA}(\text{LN}(\text{z}^l))+\text{z}^l,\\
		&	\text{z}^{l+1} = \text{MLP}(\text{LN}(\hat{\text{z}}^{l+1}))+\hat{\text{z}}^{l+1},
	\end{split}
\end{equation}
where $\hat{\text{z}}^l$ and $\text{z}^l$ denote the output features of the (S)W-MSA module and the MLP module for block $l$, respectively. 
\begin{figure*}[t]
	\centering % 图片居中
	\includegraphics[scale=0.7]{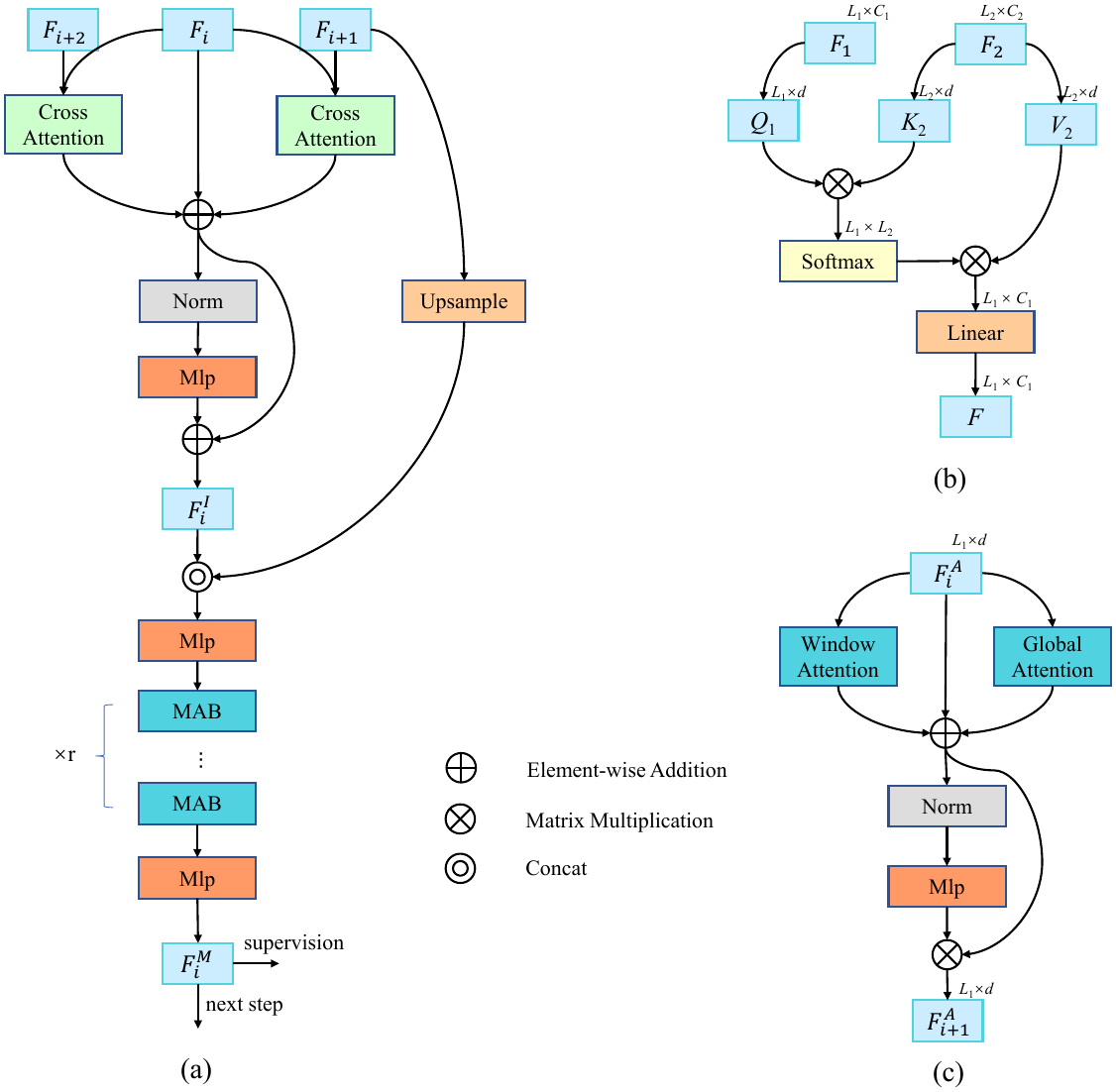}
	\caption{Details of our multistage decoder. (a) denotes one stage of the decoder, (b) denotes the structure of cross-attention, and (c) denotes the structure of the Mixed Attention Block. $r$ denotes the number of MAB stacked and we set it to 2. $F_{i+1}$ can be from $F^{I}_{i+1}$ or $F^{M}_{i+1}$. }
	\label{fig:components}
	%\vspace{-4mm}
\end{figure*}
\begin{figure}[t]
	\centering % 图片居中
	\begin{minipage}[t]{0.11\textwidth}
		\centering
		\includegraphics[scale=0.48]{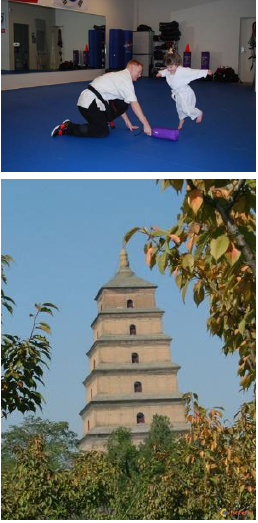}
		\centering\footnotesize{Image}
	\end{minipage}
	\begin{minipage}[t]{0.11\textwidth}
		\centering
		\includegraphics[scale=0.48]{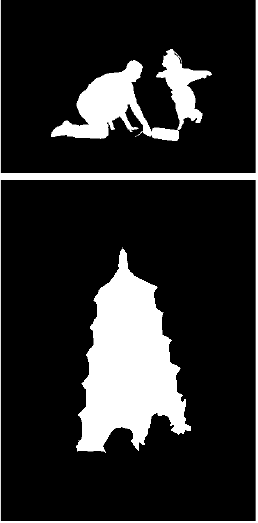}
		\centering\footnotesize{GT}
	\end{minipage}
	\begin{minipage}[t]{0.11\textwidth}
		\centering
		\includegraphics[scale=0.48]{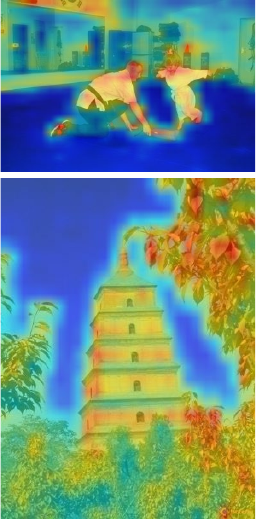}
		\centering\footnotesize{Before}
	\end{minipage}
	\begin{minipage}[t]{0.11\textwidth}
		\centering
		\includegraphics[scale=0.48]{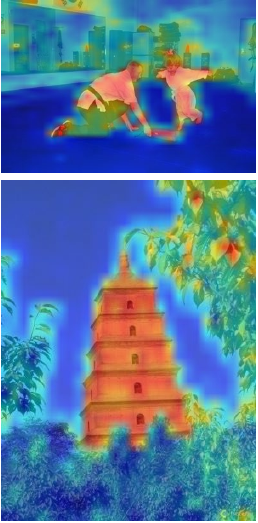}
		\centering\footnotesize{After}
	\end{minipage}
	\begin{minipage}[t]{0.1\textwidth}\end{minipage}
	
	\par\;\caption{Visual comparison of our MIB where `Before' denotes the features before MIB, and `After' denotes the features after processed. The salient area has been observably enhanced, and meanwhile, local noises have also been effectively reduced.}
	\label{fig:MIB}
	%\vspace{-4mm}
\end{figure}

\subsection{Multistage Decoder}
Figure \ref{fig:components} shows the detail of our multistage decoder. As shown in Figure \ref{fig:components} (a), in one stage, we first let multilevel features interact across 2 scales to enhance the quality of low-level features. Then, we transform multilevel features to the same resolution by upsampling method. To better integrate salient information after feature fusion, we further perform mixed attention. 

\subsubsection{Multilevel Interaction Block}
We argue that low-level features contain more non-salient information and noise, and using an all-pass skip-layer structure to concatenate the low-level features of the encoder to the decoder may cause a negative impact on the saliency map. However, low-level features are rich in local spatial details, which is crucial to gain high quality saliency maps. Some methods~\cite {MiNet,ICON} rescale the adjacent levels of features and employ upsampling or downsampling operations to align them at the same spatial resolution, followed by concatenation for further processing. Nevertheless, the non-learnable nature of commonly used sampling operations (e.g., bilinear) leads to information loss, thereby limiting the model's performance. To solve the loss of information during sampling, we propose our MIB block, which introduces the cross-attention mechanism to achieve the interaction of multilevel features, letting high-level features guide low-level features to strengthen salient regions and eliminate non-salient information. Through cross-attention, we can accomplish feature interaction across multiple scales without changing them in spatial resolution. 

In the MIB block, given the features in a sequence form $F_1 \in \mathbb{R}^{l_1 \times c_1}$, $F_2 \in \mathbb{R}^{l_2 \times c_2}$, where $F_1$ denotes low-level features and $F_2$ denotes high-level features. We first change their channel dimension and embed them to queries $Q_1 \in \mathbb{R}^{l_1 \times d}$, keys $K_2 \in \mathbb{R}^{l_2 \times d}$, and values $V_2 \in \mathbb{R}^{l_2 \times d}$ through three linear projections. Then, we compute attention between the queries from low-level features with the keys from high-level features. The output is computed as a weighted sum of the values, formulated as: 
\begin{equation}
	\small
	\begin{split}
		\text{Attention}(Q_1,K_2,V_2)=\text{Softmax}(Q_1K_2^T/\sqrt{d})V_2.
	\end{split}
\end{equation}

As shown in Figure \ref{fig:components} (a), our MIB makes features interact across two scales. Two cross-attentions are performed in parallel and gathered by element-wise addition. We follow the standard Transformer architecture~\cite{Attentionisallyouneed} in subsequent structure, including layer normalization~\cite{LN}, residual connections, formulated as:
\begin{equation}
	\begin{split}
		&\hat{F}_i = F_i+\text{CA}(F_i,F_{i+1})+\text{CA}(F_i,F_{i+2}),\\
		&F_i^I = \text{MLP}(\text{LN}(\hat{F}_i))+\hat{F}_i,
	\end{split}
\end{equation}
where $F_{i+1}$, $F_{i+2}$ denote the high-level features. $\hat{F}_i$, $F_i^I$ denote the output features of the cross-attention (CA) modules and the MLP module in the MIB, respectively. 

By using the proposed MIB, the salient region at the global level can be effectively enhanced. As can be seen in Figure \ref{fig:MIB}, after feeding the features into our MIB, the salient region is noticeably distinguished from the non-salient areas. 
%Note that, although ~\cite{VST} adopted the cross-attention mechanism to propagate long-range cross-modal interactions between RGB and depth cues, cross-attention between multilevel features still remains understudied.

\subsubsection{Mixed Attention Block}

To facilitate the aggregation of multilevel features in the sequence form, ~\cite{VST} adopts one Transformer layer with global self-attention, which may neglect local-level details and limit the fineness of the final prediction. Inspired by the success of window self-attention~\cite{Swin}, which computes self-attention within local windows, we combine global self-attention and window self-attention, aiming to model context at both global and local levels, further improve the local accuracy of the prediction map. We use the same window size (set to $7\times 7$ by default) as the encoder. 

As can be seen in Figure \ref{fig:components} (a), given feature in a sequence form $F_i^C \in \mathbb{R}^{l \times c}$ after feature fusion, we first increase its channel dimension to $d=384$ by an MLP, and then let it pass $r$ mixed attention blocks. Similar to our MIB, window self-attention and global self-attention are performed in parallel, and gathered by element-wise addition, followed by standard Transformer architecture. Finally, we apply another MLP to restore the channel dimension of the feature for subsequent processing and supervision. The whole process can be formulated as:
\begin{equation}
	\begin{split}
		&F_0^A = \text{MLP}_1(F_i^I),\\
		&\hat{F}_1^A =\text{W-MSA}(F_0^A)+\text{MSA}(F_0^A),\\
		&F_1^A = \text{MLP}(\text{LN}(\hat{F}_1^A))+\hat{F}_1^A,\\
		&\qquad...\\
		&F_i^M = \text{MLP}_2(F_r^A),\\
	\end{split}
\end{equation}
where $\hat{F}_i^A$ and $F_i^A$ denote the output features of the mixed attention module and the MLP module in our MABs, respectively. $r$ denotes the number of MAB stacked and we set it to 2. 

\subsubsection{Upsampling Methods and Multilevel Fusion}
\begin{table*}
	\centering
	\scriptsize
	\caption{Quantitative comparison of our proposed M$^3$Net with other 16 SOTA SOD methods on six benchmark datasets. The symbols “↑”/“↓” mean that a higher/lower score is better. The best results are shown in \textbf{bold}. `-R', '-R2', and `-S' means the ResNet50, Res2Net50\cite{gao2019res2net}, and SwinTransformer backbone. }
	\setlength\tabcolsep{0mm}
	\begin{tabular}{l|cccc|cccc|cccc|cccc|cccc|cccc}% 
		\hline %[2pt]  
		{dataset} & \multicolumn{4}{c|}{DUT-O} & \multicolumn{4}{c|}{DUTS} & \multicolumn{4}{c|}{ECSSD} & \multicolumn{4}{c|}{HKU-IS} & \multicolumn{4}{c|}{PASCAL-S} & \multicolumn{4}{c}{SOD}\\
		\hline %[2pt]  
		Method & $M\downarrow\:$ & $E_\xi^{m}\uparrow\:$ & $S_m\uparrow\:$ & $F_\beta^w\uparrow\:$ & $M\downarrow\:$ & $E_\xi^{m}\uparrow\:$ & $S_m\uparrow\:$ & $F_\beta^w\uparrow\:$ & $M\downarrow\:$ & $E_\xi^{m}\uparrow\:$ & $S_m\uparrow\:$ & $F_\beta^w\uparrow\:$ & $M\downarrow\:$ & $E_\xi^{m}\uparrow\:$ & $S_m\uparrow\:$ & $F_\beta^w\uparrow\:$ & $M\downarrow\:$ & $E_\xi^{m}\uparrow\:$ & $S_m\uparrow\:$ & $F_\beta^w\uparrow\:$ & $M\downarrow\:$ & $E_\xi^{m}\uparrow\:$ & $S_m\uparrow\:$ & $F_\beta^w\uparrow\:$ \\
		\hline %[2pt]  
		\multicolumn{25}{c}{CNN Methods}\\
		\hline %[2pt]  
		PiCANet & .054 & .865 & .826 & .743 & .04 & .915 & .863 & .812 & .035 & .953 & .916 & .908 & .031 & .951 & .905 & .89 & .064 & .9 & .846 & .811 & .094 & .846 & .78 & .741  \\
		%RAS & - & - & .062 & .86 & .814 & .695 & .059 & .889 & .839 & .74 & .056 & .931 & .893 & .857 & .045 & .94 & .887 & .843 & .101 & .851 & .799 & .731 & .123 & .83 & .767 & .718 \\
		BASNet & .056 & .871 & .836 & .751 & .048 & .903 & .866 & .803 & .037 & .951 & .916 & .904 & .032 & .951 & .909 & .889 & .076 & .886 & .838 & .793 & .112 & .832 & .772 & .728 \\
		CPD-R & .056 & .868 & .825 & .719 & .043 & .914 & .869 & .795 & .037 & .951 & .918 & .898 & .034 & .95 & .905 & .875 & .071 & .891 & .848 & .794 & .11 & .849 & .771 & .713 \\
		PoolNet & .054 & .867 & .831 & .725 & .037 & .926 & .887 & .817 & .035 & .956 & .926 & .904 & .03 & .958 & .919 & .888 & .065 & .907 & .865 & .809 & .103 & .867 & .792 & .746 \\
		AFNet & .057 & .861 & .826 & .717 & .046 & .91 & .867 & .785 & .042 & .947 & .913 & .886 & .036 & .949 & .905 & .869 & .07 & .895 & .849 & .797 & .108 & .847 & .78 & .726 \\
		%TSPOANet & - & - & .061 & .858 & .818 & .697 & .049 & .907 & .86 & .767 & .046 & .942 & .907 & .876 & .038 & .95 & .902 & .862 & .077 & .886 & .842 & .775 & .115 &  .844 & .775 & .718\\
		EGNet & .053 & .878 & .841 & .738 & .039 & .927 & .887 & .816 & .037 & .955 & .925 & .903 & .031 & .958 & .918 & .887 & .074 & .892 & .852 & .795 & .097 & .873 & .807 & .767 \\
		%SCRN & .056 & .875 & .837 & .72 & .04 & .925 & .885 & .803 & .037 & .956 & .927 & .9 & .034 & .956 & .916 & .876 & .063 & .91 & \textbf{.869} & .807 & -&-&-&- \\
		%F$^3$Net & {.053} & .872 & .838 & .747 & {.035} & .927 & .888 & .835 & .033 & .955 & .924 & .912 & .028 & .958 & .917 & .9 & .061 & .904 & .861 & .816 & -&-&-&-\\
		ITSD-R & .061 & .88 & .84 & .75 & .041 & .929 & .885 & .824 & .034 & .959 & .925 & .91 & .031 & .96 & .917 & .894 & .066 & .908 & .859 & .812 & .093 & .873 & .809 & .777 \\
		MINet-R & .056 & .869 & .833 & .738 & .037 & .927 & .884 & .825 & .033 & .957 & .925 & .911 & .029 & .96 & .919 & .897 & .064 & .903 & .856 & .809 & .092 & .87 & .805 & .768 \\
		LDF & \textbf{.052} & .869 & .839 & .752 & \textbf{.034} & .93 & .892 & {.845} & .034 & .954 & .924 & .915 & .028 & .958 & .919 & .904 & \textbf{.06} & .908 & .863 & .822 & .093 & .866 & .8 & .765 \\
		CSF-R2 & {.055} & .775 & .838 & .733 & .037 & .93 & .89 & .823 & .033 & .96 & .93 & .91 & .03 & .96 & .921 & .891 & .069 & .899 & .862 & .807 & .098 & {.872} & .8 & .757 \\
		GateNet-R & {.055} & .876 & {.838} & .729 & {.04} & {.928} & {.885} & .809 & .04 & .952 & .92 & .894 & .033 & .955 & .915 & .88 & .067 & .904 & .858 & .797 & .098 & .87 & .801 & .753 \\
		PFSNet & .055 & .878 & .842 & .756 & .036 & .931 & .892 & .842 & .031 & .958 & .93 & \textbf{.92} & \textbf{.026} & .962 & .924 & .91 & .063 & .906 & .86 & .819 & .089 & .875 & .81 & .781\\
		ICON-R & .057 & \textbf{.884} & {.844} & {.761} & .037 & {.932} & .889 & .837 & .032 & .96 & .929 & .918 & .029 & .96 & .92 & .902 & .064 & .908 & .861 & .818 & \textbf{.084} & \textbf{.882} & \textbf{.824} & \textbf{.794} \\
		
		M$^3$Net-R & .061 & .88 & \textbf{.848} & \textbf{.769} & .036 & \textbf{.937} & \textbf{.897} & \textbf{.849} & \textbf{.029} & \textbf{.962} & \textbf{.931} & {.919} & \textbf{.026} & \textbf{.966} & \textbf{.929} & \textbf{.913} & \textbf{.06} & \textbf{.912} & \textbf{.868} & \textbf{.827} & \textbf{.084} & .865 & .819 & .784\\
		
		\hline %[2pt]  
		\multicolumn{25}{c}{Transformer Methods}\\
		\hline %[2pt]  
		
		VST & .058 & .89 & .852 & .758 & .037 & .94 & .897 & .831 & .032 & .965 & .934 & .911 & .029 & .968 & .928 & .897 & .06 & .919 & .874 & .821 & .085 & {.876} & .82 & .776 \\
		ICON-S & \textbf{.043} & \textbf{.906} & {.869} & .804 & {.025} & \textbf{.96} & .917 & .886 & .023 & .972 & .941 & .936 & .022 & .974 & .935 & .925 & .048 & .93 & {.885} & .854 & .083 & \textbf{.885} & .825 & .802 \\ 
		SRformer & \textbf{.043} & {.894} & {.86} & .784 & {.027} & {.952} & .91 & .872 & .028 & .961 & .932 & .922 & .025 & .966 & .928 & .912 & {.051} & {.924} & {.878} & {.845} & .088 & .862 & .809 & .77 \\ 
		BBRF & .044 & {.899} & {.861} & {.803} & {.025} & {.952} & {.909} & {.886} & {.022} & {.972} & {.939} & {.944} & {.02} & {.972} & {.932} & {.932} & {.049} &  {.927} & {.878} & {.856} & .078 & .868 & .822 & .802\\
		M$^3$Net-S & .045 & .903 & \textbf{.872} & \textbf{.811} & \textbf{.024} & \textbf{.96} & \textbf{.927} & \textbf{.902} & \textbf{.021} & \textbf{.974} & \textbf{.948} & \textbf{.947} & \textbf{.019} & \textbf{.977} & \textbf{.943} & \textbf{.937} & \textbf{.047} & \textbf{.932} & \textbf{.889} & \textbf{.864} & \textbf{.073} & {.871} & \textbf{.838} & \textbf{.819} \\
		\hline %[2pt]  
	\end{tabular}
	\label{tab:Quantitative comparison}
	\vspace{-3mm}
\end{table*}
Most CNN-based methods~\cite{ICON, MiNet} adopt bilinear interpolation to re-scale high resolution feature maps, while upsampling features in the sequence form remain under-studied. In this work, we adopt the RT2T method developed in ~\cite{VST,Yuan_2021_ICCV} for token upsampling, where we found that fold with overlap leads to other upsampling methods not only in evaluation indicators but also in the visual quality of saliency map. 
%We make a comparison of different upsampling methods in M$^3$Net, and the details can be found in Section 4.

We first use a linear projection to expand the channel dimension of $F_i^I \in \mathbb{R}^{l \times c}$ to $ck^2$. Each token is seen as a $k\times k$ image patch and neighboring patches have $k-s$ overlapping, where $s$ denotes the stride. Then we fold the tokens with $p$ zero-padding. The arguments need to satisfy: 
\begin{equation}
	\small 
	\begin{split}
		l=h\times w = \lfloor\frac{h_o+2p-k}{s}+1\rfloor \times \lfloor\frac{w_o+2p-k}{s}+1\rfloor,
	\end{split}
\end{equation}
where $h_o\times w_o$ denotes the spatial size of the feature maps after folding. We reshape the feature map to the tokens with the size $l_o\times c$, where $l_o=h_o\times w_o$, and define it as $F_i^U$. We follow ~\cite{VST} to set $k=[3,3,7]$, $s=[2,2,4]$, and $p=[1,1,2]$ to get the same resolution as the original input image after three times upsampling. The upsampling ratio of each time is equal to the stride. 

As can be seen in Figure \ref{fig:components} (a), we fuse multilevel features after upsampling high-level features, and then use our MABs to further integrate salient information at different levels. One whole stage of the decoder can be formulated as:
\begin{equation}
	\begin{split}
		&F_{i+1}^U = \text{Upsample}(F_{i+1}^M),\\
		&F_i^I = \text{MIB}(F_i,F_{i+1},F_{i+2}),\\
		&F_i^C = \text{Concat}(F_i^I,F_{i+1}^U),\\
		&F_i^M = \text{MAB}s(F_i^C),
	\end{split}
\end{equation}
where $F_{i+1}^M$ denotes the output of previous stage ($F_3$ employed in the first stage) and $F_i^C$ denotes the feature maps after fusing. $F_i^M$ will be used for the next stage of the decoder and the multilevel supervision. 
\subsubsection{Multilevel Supervision Strategy}

\begin{table*}
	\scriptsize
	\centering
	\setlength\tabcolsep{1mm}
	\caption{Comparison of our proposed model with other SOTA CNN-based SOD methods on SOC test set. The best results are shown in \textbf{bold}. }
\begin{tabular}{l|r|ccccccccccccccccc|c}% 
	\hline %[2pt]  
	{Attr} & Metrics & Amulet & DSS & NLDF & C2SNet & SRM & R3Net & BMPM & DGRL & PiCA-R & RANet & AFNet & CPD & PoolNet & EGNet & BANet & SCRN & ICON-R & Ours-R \\
	\hline %[2pt]  
	\multirow{4}{*}{AC} & $M\downarrow\:$ & .120 & .113 & .119 & .109 & .096 & .135 & .098 & .081 & .093 & .132 & .084 & .083 & .094 & .085 & .086 & .078 & \textbf{.062} & {.071} \\
						& $E_\xi^{m}\uparrow\:$  & .791 & .788 & .784 & .807 & .824 & .753 & .815 & .853 & .815 & .765 & .852 & .843 & .846 & .854 & .858 & .849 & \textbf{.891} & {.885} \\
						& $S_m\uparrow\:$ & .752 & .753 & .737 & .755 & .791 & .713 & .780 & .790 & .792 & .708 & .796 & .799 & .795 & .806 & .806 & .809 & \textbf{.835} & {.824} \\
						& $F_\beta^w\uparrow\:$ & .620 & .629 & .620 & .647 & .690 & .593 & .680 & .718 & .682 & .603 & .712 & .727 & .713 & .731 & .740 & .724 & \textbf{.784} & .764 \\
	\hline %[2pt]  
	\multirow{4}{*}{BO} & $M\downarrow\:$ &.346 & .356 & .354 & .267 & .306 & .445 & .303 & .215 & .200 & .454 & .245 & .257 & .353 & .373 & .271 & .224 & .200 & \textbf{.199} \\
						& $E_\xi^{m}\uparrow\:$  & .551 & .537 & .539 & .661 & .616 & .419 & .620 & .725 & .741 & .404 & .698 & .665 & .554 & .528 & .650 & .706 & .740 & \textbf{.763} \\
						& $S_m\uparrow\:$ &.574 & .561 & .568 & .654 & .614 & .437 & .604 & .684 & .729 & .421 & .658 & .647 & .561 & .528 & .645 & .698 & \textbf{.714} & {.704} \\
						& $F_\beta^w\uparrow\:$ & .612 & .614 & .622 & .730 & .667 & .456 & .670 & .786 & .799 & .453 & .741 & .739 & .610 & .585 & .720 & .778 & \textbf{.794} & {.783} \\
	\hline %[2pt]  
	\multirow{4}{*}{CL} & $M\downarrow\:$ & .141 & .153 & .159 & .144 & .134 & .182 & .123 & .119 & .123 & .188 & .119 & .114 & .134 & .139 & .117 & .113 & .113 & \textbf{.112} \\
						& $E_\xi^{m}\uparrow\:$  & .789 & .763 & .764 & .789 & .793 & .710 & .801 & .824 & .794 & .715 & .802 & .821 & .801 & .790 & .824 & .820 & .829 & \textbf{.832} \\
						& $S_m\uparrow\:$ & .763 & .722 & .713 & .742 & .759 & .659 & .761 & .770 & .787 & .624 & .768 & .773 & .760 & .757 & .784 & \textbf{.795} & .789 & {.791} \\
						& $F_\beta^w\uparrow\:$ & .663 & .617 & .614 & .655 & .665 & .546 & .678 & .714 & .692 & .542 & .696 & .724 & .681 & .677 & .726 & .717 & .732 & \textbf{.736} \\
	\hline %[2pt]  
	\multirow{4}{*}{HO} & $M\downarrow\:$ & .119 & .124 & .126 & .123 & .115 & .136 & .116 & .104 & .108 & .143 & .103 & .097 & .100 & .106 & .094 & .096 & .092 & \textbf{.091} \\
						& $E_\xi^{m}\uparrow\:$  &.810 & .796 & .798 & .805 & .819 & .782 & .813 & .833 & .819 & .777 & .834 & .845 & .846 & .829 & .850 & .842 & .852 & \textbf{.865} \\
						& $S_m\uparrow\:$ & .791 & .767 & .755 & .768 & .794 & .740 & .781 & .791 & .809 & .713 & .798 & .803 & .815 & .802 & .819 & \textbf{.823} & .818 & \textbf{.823} \\
						& $F_\beta^w\uparrow\:$ & ..688 & .660 & .661 & .668 & .696 & .633 & .684 & .722 & .704 & .626 & .722 & .751 & .739 & .720 & .754 & .743 & .752 & \textbf{.767} \\
	\hline %[2pt]  
	\multirow{4}{*}{MB} & $M\downarrow\:$ & .142 & .132 & .138 & .128 & .115 & .160 & .105 & .113 & .099 & .139 & .111 & .106 & .121 & .109 & .104 & .100 & .100 & \textbf{.077} \\
						& $E_\xi^{m}\uparrow\:$  & .739 & .753 & .740 & .778 & .778 & .697 & .812 & .823 & .813 & .761 & .762 & .804 & .779 & .789 & .803 & .817 & .828 & \textbf{.88} \\
						& $S_m\uparrow\:$ & .712 & .719 & .685 & .720 & .742 & .657 & .762 & .744 & .775 & .696 & .734 & .754 & .751 & .762 & .764 & .792 & .774 & \textbf{.819} \\
						& $F_\beta^w\uparrow\:$ & .561 & .577 & .551 & .593 & .619 & .489 & .651 & .655 & .637 & .576 & .626 & .679 & .642 & .649 & .672 & .690 & .699 & \textbf{.74} \\
	\hline %[2pt]  
	\multirow{4}{*}{OC} & $M\downarrow\:$ & .143 & .144 & .149 & .130 & .129 & .168 & .119 & .116 & .119 & .169 & .109 & .115 & .119 & .121 & .112 & .111 & .106 & \textbf{.99} \\
						& $E_\xi^{m}\uparrow\:$  & .763 & .760 & .755 & .784 & .780 & .706 & .800 & .808 & .784 & .718 & .820 & .810 & .801 & .798 & .809 & .800 & .817 & \textbf{.843} \\
						& $S_m\uparrow\:$ & .735 & .718 & .709 & .738 & .749 & .653 & .752 & .747 & .765 & .641 & .771 & .750 & .756 & .754 & .765 & .775 & .771 & \textbf{.786} \\
						& $F_\beta^w\uparrow\:$ &.607 & .595 & .593 & .622 & .630 & .520 & .644 & .659 & .638 & .527 & .680 & .672 & .659 & .658 & .678 & .673 & .683 & \textbf{.709} \\
	\hline %[2pt]  
	\multirow{4}{*}{OV} & $M\downarrow\:$ & .173 & .180 & .184 & .159 & .150 & .216 & .136 & .125 & .127 & .217 & .129 & .134 & .148 & .146 & .119 & .126 & .120 & \textbf{.109} \\
						& $E_\xi^{m}\uparrow\:$  & .751 & .737 & .736 & .790 & .779 & .663 & .807 & .828 & .810 & .664 & .817 & .803 & .795 & .802 & .835 & .808 & .834 & \textbf{.838} \\
						& $S_m\uparrow\:$ & .721 & .700 & .688 & .728 & .745 & .624 & .751 & .762 & .781 & .611 & .761 & .748 & .747 & .752 & .779 & .774 & .779 & \textbf{.795} \\
						& $F_\beta^w\uparrow\:$ & .637 & .622 & .616 & .671 & .682 & .527 & .701 & .733 & .721 & .529 & .723 & .721 & .697 & .707 & .752 & .723 & .749 & \textbf{.759} \\
	\hline %[2pt]  
	\multirow{4}{*}{SC} & $M\downarrow\:$ &.098 & .098 & .101 & .100 & .090 & .114 & .081 & .087 & .093 & .110 & .076 & .080 & \textbf{.075} & .083 & .078 & .078 & .080 & .078 \\
						& $E_\xi^{m}\uparrow\:$  & .794 & .799 & .788 & .806 & .814 & .765 & .841 & .837 & .799 & .792 & .854 & .858 & .856 & .844 & .851 & .843 & .860 & \textbf{.871} \\
						& $S_m\uparrow\:$ & .768 & .761 & .745 & .756 & .783 & .716 & .799 & .772 & .784 & .724 & .808 & .793 & .807 & .793 & .807 & \textbf{.809} & .803 & .808 \\
						& $F_\beta^w\uparrow\:$ & .608 & .599 & .593 & .611 & .638 & .550 & .677 & .669 & .627 & .594 & .696 & .708 & .695 & .678 & .706 & .691 & \textbf{.714} & \textbf{.714} \\
	\hline %[2pt]  
	\multirow{4}{*}{SO} & $M\downarrow\:$ & .119 & .109 & .115 & .116 & .099 & .118 & .096 & .092 & .095 & .113 & .089 & .091 & .087 & .098 & .090 & .082 & .087 & \textbf{.079} \\
						& $E_\xi^{m}\uparrow\:$  & .745 & .756 & .747 & .752 & .769 & .732 & .780 & .802 & .766 & .759 & .792 & .804 & .814 & .784 & .801 & .797 & .816 & \textbf{.835} \\
						& $S_m\uparrow\:$ & .718 & .713 & .703 & .706 & .737 & .682 & .732 & .736 & .748 & .682 & .746 & .745 & .768 & .749 & .755 & .767 & .763 & \textbf{.778} \\
						& $F_\beta^w\uparrow\:$ & .523 & .524 & .526 & .531 & .561 & .487 & .567 & .602 & .566 & .518 & .596 & .623 & .626 & .594 & .621 & .614 & .634 & \textbf{.66} \\
	\hline    
\end{tabular}
\label{tab:soc}
\end{table*}

After each stage of our multistage decoder, the channel of features will be reduced to 1-dimension, expressed as $F_i^P \in \mathbb{R}^{l \times 1}$, for multilevel supervision. Previous methods mostly apply BCE as the loss function, which treats every pixel equally and may neglect the relation between pixels. We additionally use the IoU loss~\cite{BASNet,8237634} to supervise the structural inconsistency between the prediction and the ground truth. The BCE loss is defined as:
\begin{equation}
	\small
	\begin{split}
		\mathcal{L}_{BCE}=-\sum_{x=1}^{H}\sum_{y=1}^{W}[G(x,y)logP(x,y)\\+(1-G(x,y))log(1-P(x,y))],
	\end{split}
\end{equation}
where $H$, $W$ are the height and width of the image, and $P(x,y)$, $G(x,y)$ denote the pixels of the prediction and the ground truth at position $(x,y)$, respectively. Meanwhile, the IoU loss is formulated as:
\begin{equation}
	\small
	\begin{split}
		\mathcal{L}_{IoU}=1-\frac{\sum_{x=1}^{H}\sum_{y=1}^{W}(P(x,y)G(x,y))}{\sum_{x=1}^{H}\sum_{y=1}^{W}(P(x,y)+G(x,y)-P(x,y)G(x,y))},
	\end{split}
\end{equation}
Then the joint loss can be defined as:
\begin{equation}
	\small
	\begin{split}
		\mathcal{L}(P,G)=\mathcal{L}_{BCE}+\mathcal{L}_{IoU}.
	\end{split}
\end{equation}
During training, we also use the multilevel supervision strategy widely used in \cite{PoolNet,MiNet,F3Net,GCPANet,VST} to facilitate the training process. %and allocate different weights to them. The total loss is defined as:
%\begin{equation}
%\small
%\begin{split}
%	\mathcal{L}_{total}=1\times\mathcal{L}(P_0,G_0)+0.8\times\mathcal{L}(P_1,G_1)\\+0.5\times\mathcal{L}(P_2,G_2)
%\end{split}
%\end{equation}
%where $P_i$, $G_i$ denote the prediction and the ground truth from different levels.

\section{Experiments}

\subsection{Implementation Details}
For fair comparison, we follow recent methods to use the \textbf{DUTS-TR} (10553 images)~\cite{DUTS} to train our M$^3$Net and we resize images to 224×224. Normalization, random $90^\circ$ rotation, and crop were applied as the data augmentation. According to ~\cite{BigBatchsize}, we set the batch size as 8 to avoid weakening the generalization ability of the model. We use the Adam optimizer to train our network for 120 epochs and set its initial learning rate to 0.0001. In order to make the model converge better, we reduce the learning rate to 0.00002 in the last 20 epochs. All experiments were implemented on an RTX 3090 GPU.
\subsection{Evaluation Datasets and Metrics}

We evaluate our M$^3$Net on six widely used benchmark datasets, including \textbf{DUT-OMORN}~\cite{DUT-O} (5168 images), \textbf{DUTS-TE}~\cite{DUTS} (5019 images), \textbf{ECSSD}~\cite{ECSSD} (1000 images), \textbf{HKU-IS}~\cite{HKU-IS} (4447 images), \textbf{PASCAL-S}~\cite{PASCAL-S} (850 images), \textbf{SOD}~\cite{SOD} (300 images). 

We use four metrics to evaluate our model and the existing state-of-the-art algorithms:

\textbf{(1) MAE}. MAE evaluates the average pixel-wise difference between the prediction $P$ and the ground-truth $G$, and is computed as $MAE = \frac{1}{H\times W}\sum_{x=1}^{H}\sum_{y=1}^{W}\vert P(x,y)-G(x,y)\vert$.

%\textbf{(2) F-measure}. F-measure\cite{Achanta2009FrequencytunedSR} considers both precision and recall value of the prediction map and can be computed as $F_\beta=\frac{(1+\beta^2)\times Precision\times Recall}{\beta^2\times Precision+Recall}$, where $\beta^2$ is set to 0.3 to put more emphasis on precision. 

\textbf{(2) E-measure}. E-measure~\cite{ijcai2018p97} considers the local pixel values with the image-level mean value simultaneously: $E_\xi=\frac{1}{H\times W}\sum_{x=1}^{H}\sum_{y=1}^{W}\phi_\xi(i,j)$ where $\phi_\xi$ denotes the enhanced alignment matrix. 

\textbf{(3) S-measure}. S-measure~\cite{fan2017structure} aims to measure the structural similarity between the prediction and ground truth, which is computed as $S_m = \alpha S_o +(1-\alpha)S_r$. $S_o$ and $S_r$ denote the object-level and region-level structural similarity, respectively, and $\alpha$ is set to 0.5. 

\textbf{(4) Weighted F-measure}. Weighted F-measure~\cite{6909433} gives an intuitive generalization of $F_\beta$, calculated as $F_\beta^w=\frac{(1+\beta^2)\times Precision^w\times Recall^w}{\beta^2\times Precision^w+Recall^w}$. It extends the four basic quantities TP, TN, FP and FN to real values, and assigns different weights $(w)$ to different errors based on different locations and neighborhood information.
\subsection{Comparison with State-of-the-Art}
\begin{table*}
	\centering
	\scriptsize
	\caption{The performance of our model under different inputs and backbones with varying scales. }
	\label{tab:size}
	\setlength\tabcolsep{1.5mm}
	\begin{tabular}{c|c|c|c|c|cccc|cccc|cccc}% 
		\hline
		\multirow{2}{*}{Method} & {\multirow{2}{*}{Backbone}} & {\multirow{2}{*}{Input Size}} & {\multirow{2}{*}{MACs}} & {\multirow{2}{*}{Params}} & \multicolumn{4}{c|}{DUTS} & \multicolumn{4}{c|}{ECSSD} & \multicolumn{4}{c}{HKU-IS}\\
		
		& &&&& $M\downarrow\:$ & $E_\xi^{m}\uparrow\:$ & $S_m\uparrow\:$ & $F_\beta^w\uparrow\:$ & $M\downarrow\:$ & $E_\xi^{m}\uparrow\:$ & $S_m\uparrow\:$ & $F_\beta^w\uparrow\:$ & $M\downarrow\:$ & $E_\xi^{m}\uparrow\:$ & $S_m\uparrow\:$ & $F_\beta^w\uparrow\:$ \\
		\hline %[2pt] 
		Ours-R & ResNet50 & 224$\times$224 & 18.83 & 34.61 & 0.036 & {0.937} & {0.897} & {0.849} & {0.029} & {0.962} & {0.931} & {0.919} & {0.026} & {0.966} & {0.929} & {0.913} \\
		Ours-R & ResNet50 & 352$\times$352 & 46.49 & 34.61 & 0.039 & 0.931 & 0.898 & 0.856 & 0.03 & 0.959 & 0.932 & 0.923 & 0.025 & 0.964 & 0.93 & 0.916\\
		Ours-S & SwinS & 224$\times$224 & 23.11 & 59.07 & 0.028 & 0.952 & 0.913 & 0.878 & 0.024 & 0.97 & 0.941 & 0.935 & 0.023 & 0.97 & 0.934 & 0.921\\
		Ours-S & SwinB & 224$\times$224 & 39.51 & 102.62 & 0.026 & 0.955 & 0.918 & 0.885 & 0.022 & 0.974 & 0.945 & 0.941 & 0.021 & 0.975 & 0.939 & 0.928\\
		Ours-S & SwinB & 384$\times$384 & 116.12 & 102.75 & {0.024} & {0.96} & {0.927} & {0.902} & {0.021} & {0.974} & {0.948} & {0.947} & {0.019} & {0.977} & {0.943} & {0.937} \\
		\hline 
	\end{tabular}
	\vspace{-3mm}
\end{table*}
\begin{figure*}[ht]
	\centering % 图片居中
	\includegraphics[scale=0.24]{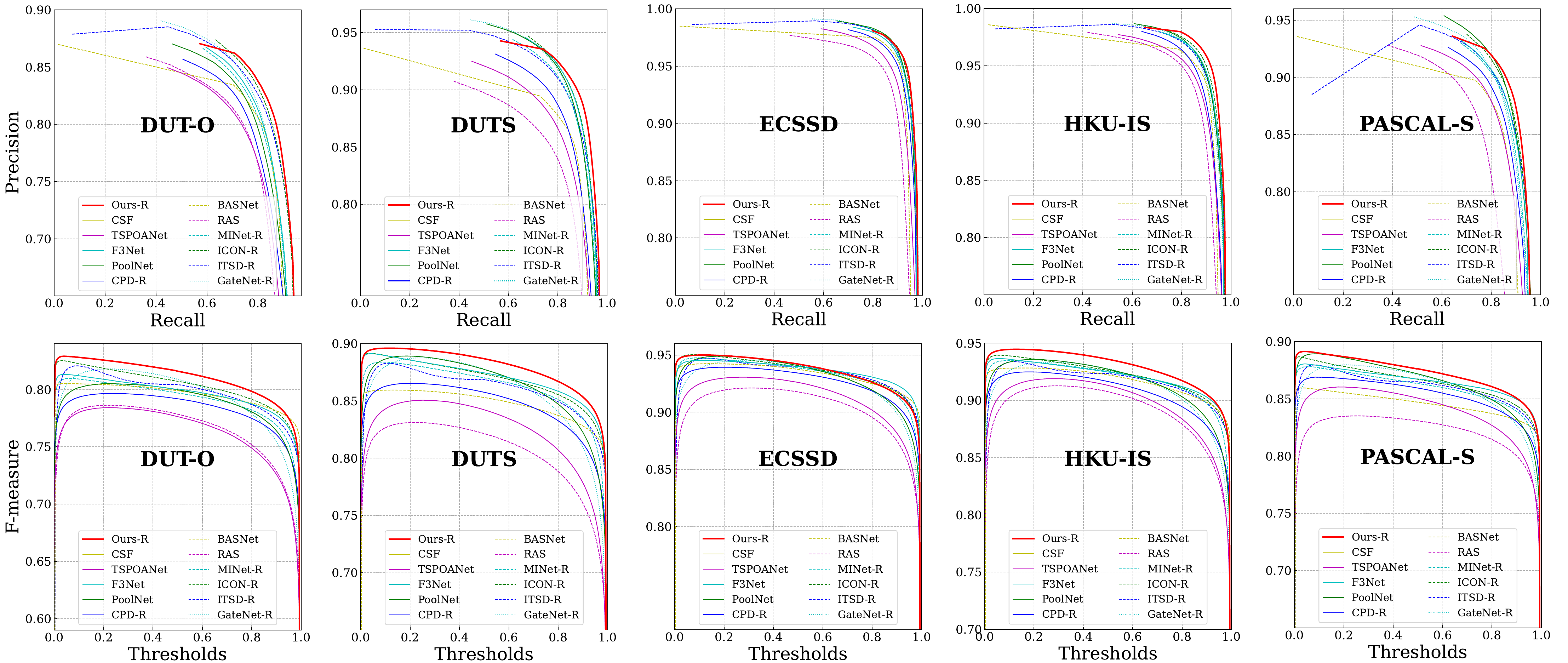}
	\caption{Precision-Recall curves (row 1) and F-measure curves (row 2) of our M$^3$Net-R and other CNN SOTA methods on five benchmark datasets. }
	\label{fig:PR&Fm}
	%\vspace{-4mm}
\end{figure*}

\begin{figure*}
	\centering % 图片居中
	\begin{minipage}[t]{0.09\textwidth}
		\centering
		\includegraphics[scale=0.82]{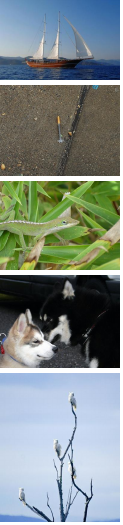}
		\centering\footnotesize{Image}
	\end{minipage}
	\begin{minipage}[t]{0.09\textwidth}
		\centering
		\includegraphics[scale=0.82]{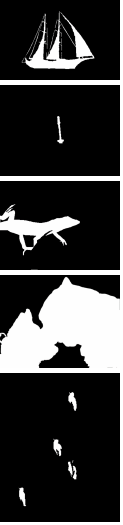}
		\centering\footnotesize{GT}
	\end{minipage}
	\begin{minipage}[t]{0.09\textwidth}
		\centering
		\includegraphics[scale=0.82]{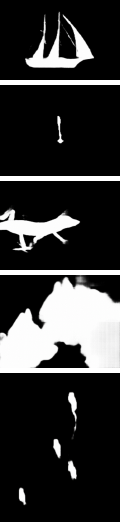}
		\centering\footnotesize{M$^3$Net-S}
	\end{minipage}
	\begin{minipage}[t]{0.09\textwidth}
		\centering
		\includegraphics[scale=0.82]{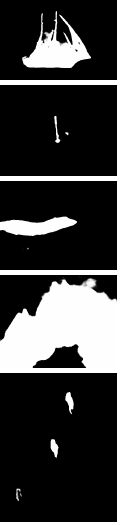}
		\centering\footnotesize{BBRF\par\cite{BBRF}}
	\end{minipage}
	\begin{minipage}[t]{0.09\textwidth}
		\centering
		\includegraphics[scale=0.82]{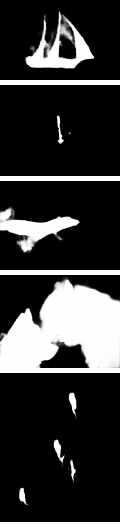}
		\centering\footnotesize{ICON-S\par\cite{ICON}}
	\end{minipage}
	\begin{minipage}[t]{0.09\textwidth}
		\centering
		\includegraphics[scale=0.82]{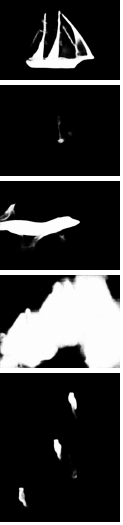}
		\centering\footnotesize{SRformer\par\cite{SelfReformer}}
	\end{minipage}
	\begin{minipage}[t]{0.09\textwidth}
		\centering
		\includegraphics[scale=0.82]{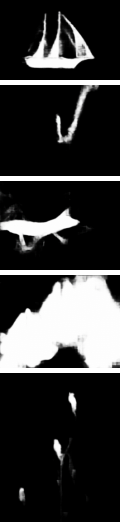}
		\centering\footnotesize{VST\par\cite{VST}}
	\end{minipage}
	\begin{minipage}[t]{0.09\textwidth}
		\centering
		\includegraphics[scale=0.82]{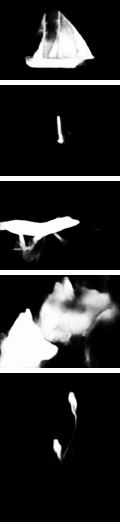}
		\centering\footnotesize{GateNet\par\cite{GateNet}}
	\end{minipage}
	\begin{minipage}[t]{0.09\textwidth}
		\centering
		\includegraphics[scale=0.82]{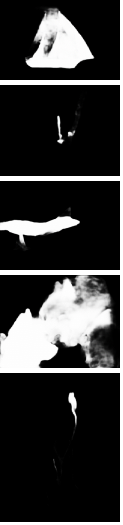}
		\centering\footnotesize{MiNet\par\cite{MiNet}}
	\end{minipage}
	\begin{minipage}[t]{0.09\textwidth}
		\centering
		\includegraphics[scale=0.82]{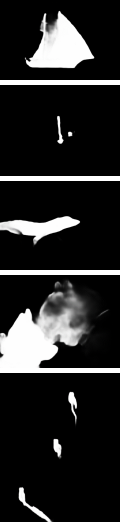}
		\centering\footnotesize{F3Net\par\cite{F3Net}}
	\end{minipage}
	\begin{minipage}[t]{0.1\textwidth}
	\end{minipage}
	\par\;\caption{Visual comparisons between our M$^3$Net and other 7 state-of-the-art methods on various scenes. 
		In contrast to previous methods, our M$^3$Net generates prediction maps with fewer shadows and undersaturated regions, thereby enhancing its reliability. }
	\label{fig:visualcomparison}
	\vspace{-3mm}
\end{figure*}

We compare our M$^3$Net with 30 state-of-the-art methods, including PiCANet~\cite{PiCANet}, RAS~\cite{RAS}, Amulet\cite{Amulet}, DSS\cite{DSS}, NLDF\cite{luo2017non}, C2SNet\cite{xin2018c2s}, SRM\cite{8237695}, R3Net\cite{R3Net}, BMPM\cite{Zhang_2018_CVPR}, DGRL\cite{Wang_2018_CVPR}, RANet\cite{8966594}, BANet\cite{Su_2019_ICCV}, BASNet~\cite{BASNet}, CPD~\cite{CPD}, PoolNet~\cite{PoolNet}, AFNet~\cite{AFNet}, TSPOANet~\cite{TSPOANet}, EGNet~\cite{EGNet}, SCRN~\cite{SCRN}, F3Net~\cite{F3Net}, ITSD~\cite{ITSD}, MiNet~\cite{MiNet}, LDF~\cite{CVPR2020_LDF}, CSF~\cite{gao2020sod100k}, GateNet~\cite{GateNet}, PFSNet\cite{ma2021pyramidal}, ICON~\cite{ICON}, 
VST~\cite{VST}, SRformer~\cite{SelfReformer} and BBRF\cite{BBRF}. 
%For a fair comparison, we generated the prediction maps of VST, SRformer, and ICON-S in the same environment, using their pre-trained weights. 
We also adopt ResNet-50 as the backbone to compare with CNN-based methods, in which case our model is referred to as M$^3$Net-R. %The detailed model architecture of M$^3$Net-R can be found in supplementary material. 

\subsubsection{Quantitative Comparison}

Table \ref{tab:Quantitative comparison} shows the quantitative comparison results on six widely used benchmark datasets. We compare our method with 16 state-of-the-art methods in terms of MAE, $E_\xi$, $S_m$, $F_\beta^w$. 
Regarding M$^3$Net-R, an input size of 224 x 224 is adopted. In the case of M$^3$Net-S, an input size of 384 x 384 is used. 
The results show that our M$^3$Net not only outperforms all previous state-of-the-art Transformer based methods but also achieves competitive results in CNN-based methods. 
We also report the performance of our model under different inputs and backbones with varying scales. 
As shown in Table \ref{tab:size}, with the increase in the size of the input image, both the performance and computational cost of the model increase simultaneously. 
Therefore, we can adjust the input size of the model to accommodate different computational requirements.
In addition, we present the precision-recall~\cite{6871397} and F-measure curves~\cite{Achanta2009FrequencytunedSR} in Figure \ref{fig:PR&Fm}. 

Recently, Fan \textit{et al.} proposed a challenging dataset known as SOC\cite{fan2022salient}, which is based on real-world scenarios. In contrast to previous datasets, the SOC dataset comprises more complex scenes and is divided into nine distinct subsets based on image attributes, including AC (appearance change), BO (big object), CL (clutter), HO (heterogeneous object), MB (motion blur), OC (occlusion), OV (out-of-view), SC (shape complexity), and SO (small object). 
Evaluating the performance of our proposed model in comparison to previous methods on the SOC dataset provides a more comprehensive validation of its performance. 
Table \ref{tab:soc} presents a comparison between our model and 17 recent state-of-the-art CNN-based methods, focusing on attribute-based performance. 
Notably, our model demonstrates significant performance improvement compared to the existing methods. 

\subsubsection{Visual Comparison}
Figure \ref{fig:visualcomparison} provides visual comparisons between our method and other methods. As can be seen, our M$^3$Net reconstructs more accurate saliency maps, while local fine-grained details are well preserved. Besides, our method excels in dealing with challenging cases like small objects (row 2), low-contrast (row 4), complex backgrounds (row 3), delicate structures (row 1), and multiple objects (rows 4 and 5) integrally and noiselessly. The above results show the versatility and robustness of our M$^3$Net.
\subsection{Ablation Studies}
\begin{table*}
	\centering
	\scriptsize
	\caption{Ablation studies of our proposed model. ``Bi" denotes bilinear upsampling, ``F" denotes fold upsampling, ``MIB" denotes our proposed multilevel interaction block, and ``MAB" denotes our proposed mixed attention block. }
	\label{tab:ablation}
	\setlength\tabcolsep{2.5mm}
	\begin{tabular}{c|l|cccc|cccc|cccc}% 
		\hline
		\multirow{2}{*}{ID} & {\multirow{2}{*}{Component Settings}} & \multicolumn{4}{c|}{DUTS} & \multicolumn{4}{c|}{ECSSD} & \multicolumn{4}{c}{HKU-IS}\\
		
		& & $M\downarrow\:$ & $E_\xi^{m}\uparrow\:$ & $S_m\uparrow\:$ & $F_\beta^w\uparrow\:$ & $M\downarrow\:$ & $E_\xi^{m}\uparrow\:$ & $S_m\uparrow\:$ & $F_\beta^w\uparrow\:$ & $M\downarrow\:$ & $E_\xi^{m}\uparrow\:$ & $S_m\uparrow\:$ & $F_\beta^w\uparrow\:$ \\
		\hline %[2pt] 
		1 & +Bi & 0.046 & 0.9 & 0.851 & 0.78 & 0.046 & 0.936 & 0.897 & 0.879 & 0.035 & 0.945 & 0.896 & 0.872\\
		2 & +F & 0.042 & 0.911 & 0.868 & 0.802 & 0.042 & 0.939 & 0.906 & 0.888 & 0.031 & 0.953 & 0.911 & 0.888\\
		3 & +F+MIB & 0.038 & 0.93 & 0.891 & 0.84 & 0.035 & 0.953 & 0.923 & 0.907 & 0.028 & 0.96 & 0.923 & 0.904\\
		4 & +F+MAB & 0.037 & 0.933 & 0.892 & 0.842 & 0.035 & 0.953 & 0.924 & 0.908 & 0.028 & 0.961 & 0.923 & 0.904\\
		5 & +F+MIB+MAB & \textbf{0.036} & \textbf{0.937} & \textbf{0.897} & \textbf{0.849} & \textbf{0.029} & \textbf{0.962} & \textbf{0.931} & \textbf{0.919} & \textbf{0.026} & \textbf{ 0.966} & \textbf{0.929} & \textbf{0.913} \\
		\hline 
	\end{tabular}
	\vspace{-3mm}
\end{table*}
\begin{table}
	\centering
	\scriptsize
	\caption{Ablation Studies of different feature enhancement methods compared with our MIB. }
	\label{tab:mib_vs_fems}
	\setlength\tabcolsep{0.5mm}
	\begin{tabular}{c|c|ccc|ccc|ccc}% 
		\hline
		\multirow{2}{*}{ID} & \multirow{2}{*}{FEMs Settings} & \multicolumn{3}{c|}{DUTS} & \multicolumn{3}{c|}{ECSSD} & \multicolumn{3}{c}{HKU-IS}\\
		&& $M\downarrow$ & $S_m\uparrow$ & $F_\beta^w\uparrow$ & $M\downarrow$ & $S_m\uparrow$ & $F_\beta^w\uparrow$ & $M\downarrow$ & $S_m\uparrow$ & $F_\beta^w\uparrow$ \\
		\hline
		3& MIB & \textbf{.038} & \textbf{.891} & \textbf{.84} & \textbf{.035} & \textbf{.923} & \textbf{.907} & \textbf{.028} & \textbf{.923} & \textbf{.904} \\
		6& DFA\cite{ICON} & .045 & .878 & .796 & .04 & .92 & .892 & .035 & .915 & .88 \\
		%3& +Inception & .044 & {.876} & {.795} & {.041} & {.917} & {.887} & {.035} & {.914} & {.876} \\
		%6& +ASPP\cite{ASPP} & .045 & {.87} & {.783} & {.046} & {.908} & {.876} & {.037} & {.908} & {.868} \\
		%7& +PSP\cite{zhao2017pspnet} & .044 & {.879} & {.797} & {.041} & {.919} & {.888} & {.036} & {.915} & {.877} \\
		7& RFB\cite{Liu_2018_ECCV} & .043 & {.882} & {.807} & {.039} & {.921} & {.893} & {.035} & {.915} & {.88} \\
		8& AIM\cite{MiNet} & .041 & {.888} & {.809} & {.037} & \textbf{.923} & {.898} & {.033} & {.919} & {.887} \\
		\hline
	\end{tabular}
\end{table}
\begin{table}
	\centering
	\scriptsize
	\caption{Different cross-levels in our MIB. }
	\label{tab:mib}
	\setlength\tabcolsep{0.5mm}
	\begin{tabular}{c|c|ccc|ccc|ccc}% 
		\hline
		\multirow{2}{*}{ID} & \multirow{2}{*}{Across Levels} & \multicolumn{3}{c|}{DUTS} & \multicolumn{3}{c|}{ECSSD} & \multicolumn{3}{c}{HKU-IS}\\
		&& $M\downarrow$ & $S_m\uparrow$ & $F_\beta^w\uparrow$ & $M\downarrow$ & $S_m\uparrow$ & $F_\beta^w\uparrow$ & $M\downarrow$ & $S_m\uparrow$ & $F_\beta^w\uparrow$ \\
		\hline
		3& 2 & \textbf{.038} & \textbf{.891} & \textbf{.84} & {.035} & \textbf{.923} & \textbf{.907} & \textbf{.028} & \textbf{.923} & \textbf{.904} \\
		9& 1 & .039 & .889 & .837 & \textbf{.034} & \textbf{.923} & .906 & .029 & {.922} & \textbf{.904} \\ 
		10& 3 & .039 & .887 & .835 & .037 & {.92} & .906 & .03 & {.922} & .903 \\
		\hline
	\end{tabular}
\end{table}
\begin{table}
	\centering
	\scriptsize
	\caption{Different interaction modes in our MIB. }
	\label{tab:mib2}
	\setlength\tabcolsep{0.5mm}
	\begin{tabular}{c|c|ccc|ccc|ccc}% 
		\hline
		\multirow{2}{*}{ID} & \multirow{2}{*}{Interaction Settings} & \multicolumn{3}{c|}{DUTS} & \multicolumn{3}{c|}{ECSSD} & \multicolumn{3}{c}{HKU-IS}\\
		&& $M\downarrow$ & $S_m\uparrow$ & $F_\beta^w\uparrow$ & $M\downarrow$ & $S_m\uparrow$ & $F_\beta^w\uparrow$ & $M\downarrow$ & $S_m\uparrow$ & $F_\beta^w\uparrow$ \\
		\hline
		3& high to low & \textbf{.038} & \textbf{.891} & \textbf{.84} & \textbf{.035} & \textbf{.923} & \textbf{.907} & \textbf{.028} & \textbf{.923} & \textbf{.904} \\
		11& low to high & .042 & .886 & .836 & {.038} & {.911} & .898 & .031 & {.914} & {.891} \\ 
		12& bi-directional & .039 & .889 & .837 & .036 & \textbf{.923} & \textbf{.907} & .029 & {.922} & .902 \\
		\hline
	\end{tabular}
\end{table}
To demonstrate the effectiveness of different modules in our M$^3$Net, we conduct the quantitative results of several simplified versions of our method. The experimental results on three datasets including \textbf{DUTS-TE}, \textbf{ECSSD}, \textbf{HKU-IS} are given in Table \ref{tab:ablation}. We start from a UNet-like structure with skip connections and bilinear upsampling as the baseline and progressively incorporate the proposed modules, including MIB and MAB. 

%\noindent\textbf{Effectiveness of multilevel features fusion.} We progressively upsample and fuse multilevel features in our decoder (ID: 3) to gather salient information from different scales. We find that low-level features bring sufficient local information for reconstructing the saliency map. 

\subsubsection{Effectiveness of MIB} 
For enhancing multilevel features from the encoder, we use our MIB to strengthen salient regions and reduce non-salient information of low-level features, shown as ``+MIB" in Table 2. The results show that our MIB gains notable improvement, demonstrating its effectiveness. 

To verify the superiority of the proposed MIB, 
we compared it with existing feature enhancement methods, 
including DFA\cite{ICON}, RFB\cite{Liu_2018_ECCV}, and AIM\cite{MiNet}. 
As can be seen in Table \ref{tab:mib_vs_fems}, 
the proposed MIB demonstrates advantages across all performance metrics. 
This suggests that utilizing the complementarity of multiscale features can lead to improved feature enhancement effects. 

To further investigate the effectiveness of our MIB and
the interaction between multilevel features, we made adjustments to the scale range of the MIB. 
For instance, when we combine two scales, we employ features at 1/8 and 1/16 levels to interact with features at the 1/4 level. 
Conversely, when we intersect two scales, we exclusively use features at the 1/8 level to interact with features at the 1/4 level. 
The results are presented in Table \ref{tab:mib}. 

It is worth mentioning that the interaction within our MIB is unidirectional, 
where the high-level features solely guide the low-level features. 
Hence, we also endeavored to integrate reverse and bidirectional interaction in our MIB, 
and the outcomes are showcased in Table \ref{tab:mib2}. 
The low-to-high interaction is observed to result in performance degradation, 
which could be attributed to the presence of noise and non-salient information in the low-level features. 

%\begin{figure}[t]
%	\centering % 图片居中
%	\includegraphics[scale=0.65]{figures/upsample methods.pdf}
%	\caption{Visual comparison of the ablation study about upsampling methods. }
%	\label{fig:upsample}
%\end{figure}
	
\subsubsection{Effectiveness of MAB}
To better integrate salient information from multilevel features after fusing them, we use our MABs to model context at both global and local levels to gain high quality salient map, shown as ``+MAB" in Table \ref{tab:ablation}. The results demonstrate the effectiveness of our MAB. 

In order to demonstrate the superior performance of the proposed MAB 
in saliency integration, we conduct a comparative evaluation with existing methods, 
including SIM\cite{MiNet}, AFM\cite{Ma_Xia_Li_2021}, shown as Table \ref{tab:mab_vs_sims}. 
The effectiveness of the proposed MAB in integrating salient information while preserving intricate local details is evident, resulting in the generation of saliency predictions of high quality.

\begin{figure}
	\centering % 图片居中
	\begin{minipage}[t]{0.088\textwidth}
		\centering
		\includegraphics[scale=0.68]{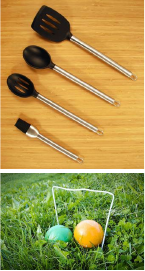}
		\centering\footnotesize{Image}
	\end{minipage}
	\begin{minipage}[t]{0.088\textwidth}
		\centering
		\includegraphics[scale=0.68]{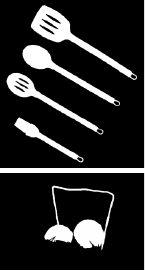}
		\centering\footnotesize{GT}
	\end{minipage}
	\begin{minipage}[t]{0.088\textwidth}
		\centering
		\includegraphics[scale=0.68]{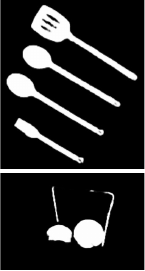}
		\centering\footnotesize{Window}
	\end{minipage}
	\begin{minipage}[t]{0.088\textwidth}
		\centering
		\includegraphics[scale=0.68]{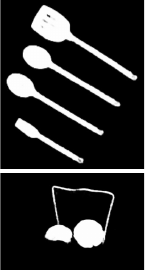}
		\centering\footnotesize{Global}
	\end{minipage}
	\begin{minipage}[t]{0.088\textwidth}
		\centering
		\includegraphics[scale=0.68]{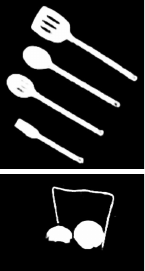}
		\centering\footnotesize{Mixed}
	\end{minipage}
	\begin{minipage}[t]{0.1\textwidth}\end{minipage}
	\par\;\caption{Visual comparison of the ablation study on our MAB. 
		It can be seen that the mixed attention, which combines window attention and global attention, can not only ensure accurate object localization but also effectively preserve local details. }
	\label{fig:MAB}
	%\vspace{-4mm}
\end{figure}

To further explore the effectiveness of mixed attention, we conduct extra experiments, the results are shown in Table \ref{tab:mab} and Figure \ref{fig:MAB}. We find that mixed attention can better retain local details with a slight additional computational cost. Besides, $7\times 7$ is also a more suitable window size. 
\begin{table}
	\centering
	\scriptsize
	\caption{Ablation studies of different saliency integration methods compared with our MAB. }
	\label{tab:mab_vs_sims}
	\setlength\tabcolsep{0.5mm}
	\begin{tabular}{c|c|ccc|ccc|ccc}% 
		\hline
		\multirow{2}{*}{ID} & \multirow{2}{*}{Integration Methods} & \multicolumn{3}{c|}{DUTS} & \multicolumn{3}{c|}{ECSSD} & \multicolumn{3}{c}{HKU-IS}\\
		&& $M\downarrow$ & $S_m\uparrow$ & $F_\beta^w\uparrow$ & $M\downarrow$ & $S_m\uparrow$ & $F_\beta^w\uparrow$ & $M\downarrow$ & $S_m\uparrow$ & $F_\beta^w\uparrow$ \\
		\hline
		5 & MAB & \textbf{.036} & \textbf{.897} & \textbf{.849} & \textbf{.029} & \textbf{.931} & \textbf{.919} & \textbf{.026} & \textbf{.929} & \textbf{.913} \\
		13 & SIM\cite{MiNet} & .037 & {.895} & {.834} & {.035} & {.925} & {.904} & {.032} & {.921} & {.893} \\
		14 & AFM\cite{Ma_Xia_Li_2021} & .04 & {.894} & {.837} & {.036} & {.923} & {.901} & {.031} & {.923} & {.894} \\
		
		%8& WA & \textbf{.028} & .908 & .868 & \textbf{.025} & .939 & \textbf{.933} & .024 & .933 & .92 \\
		\hline 
	\end{tabular}
	%\vspace{-3mm}
\end{table}
\begin{table}
	\centering
	\scriptsize
	\caption{Ablation study of different attention settings in our MAB. $h\times w$ can be seen as global self-attention. }
	\label{tab:mab}
	\setlength\tabcolsep{0.3mm}
	\begin{tabular}{c|c|ccc|ccc|ccc}% 
		\hline
		\multirow{2}{*}{ID} & \multirow{2}{*}{Attention Settings} & \multicolumn{3}{c|}{DUTS} & \multicolumn{3}{c|}{ECSSD} & \multicolumn{3}{c}{HKU-IS}\\
		&& $M\downarrow$ & $S_m\uparrow$ & $F_\beta^w\uparrow$ & $M\downarrow$ & $S_m\uparrow$ & $F_\beta^w\uparrow$ & $M\downarrow$ & $S_m\uparrow$ & $F_\beta^w\uparrow$ \\
		\hline
		15 & $h\times w$ & \textbf{.036} & .894 & .844 & {.03} & {.929} & .918 & {.027} & {.927} & .91 \\
		16 & $7\times 7$ & .038 & .895 & .843 & \textbf{.029} & {.929} & .917 & {.028} & {.927} & .908 \\
		5 & $7\times 7+h\times w$ & \textbf{.036} & \textbf{.897} & \textbf{.849} & \textbf{.029} & \textbf{.931} & \textbf{.919} & \textbf{.026} & \textbf{.929} & \textbf{.913} \\
		17 & $4\times 4+h\times w$ & {.038} & {.891} & {.839} & {.03} & {.929} & {.917} & {.028} & {.926} & {.906} \\
		18 & $14\times 14+h\times w$ & \textbf{.036} & {.896} & {.847} & \textbf{.029} & \textbf{.931} & {.918} & {.027} & {.928} & {.91} \\
		19 & $7\times 7+14\times 14$ & \textbf{.036} & .894 & .845 & {.03} & .929 & .917 & {.026} & {.926} & .909 \\
		%25& $7\times 7+14\times 14+28\times 28$ & {.036} & \textbf{.897} & {.849} & {.029} & {.931} & {.919} & {.026} & {.928} & {.911} \\
		\hline
	\end{tabular}
\end{table}
\subsubsection{Evaluation of upsampling method}
\begin{table}
	\centering
	\scriptsize
	\caption{Ablation studies of the upsampling methods adopted in the M$^3$Net. }
	\label{tab:ablation_up}
	\setlength\tabcolsep{0.3mm}
	\begin{tabular}{c|c|ccc|ccc|ccc}% 
		\hline
		\multirow{2}{*}{ID} & \multirow{2}{*}{Upsampling Methods} & \multicolumn{3}{c|}{DUTS} & \multicolumn{3}{c|}{ECSSD} & \multicolumn{3}{c}{HKU-IS}\\
		&& $M\downarrow$ & $S_m\uparrow$ & $F_\beta^w\uparrow$ & $M\downarrow$ & $S_m\uparrow$ & $F_\beta^w\uparrow$ & $M\downarrow$ & $S_m\uparrow$ & $F_\beta^w\uparrow$ \\
		\hline
		20 & pixel shuffle & \textbf{.036} & .895 & .845 & .031& .928 & .915 & .027 & .926 & .907 \\ 
		21 & bilinear & .037 & .89 & .837 & .031 & .927 & .914 & .028 & .923 & .903 \\
		%11& transposed convolution & \textbf{.029} & .909 & .87 & \textbf{.025} & .94 & .931 & .024 & .932 & .919 \\
		22 & fold & .037 & .893 & .843 & \textbf{.029} & .93 & .917 & \textbf{.026} & {.927} & .91 \\
		5 & fold with overlap & \textbf{.036} & \textbf{.897} & \textbf{.849} & \textbf{.029} & \textbf{.931} & \textbf{.919} & \textbf{.026} & \textbf{.929} & \textbf{.913} \\
		\hline
	\end{tabular}
	%\vspace{-4mm}
\end{table}
\begin{figure}
	\centering % 图片居中
	\begin{minipage}[t]{0.07\textwidth}
		\centering
		\includegraphics[scale=0.45]{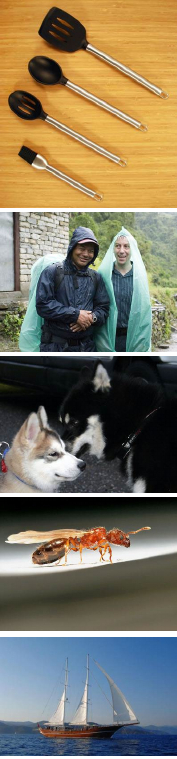}
		\centering\footnotesize{Image}
	\end{minipage}
	\begin{minipage}[t]{0.07\textwidth}
		\centering
		\includegraphics[scale=0.45]{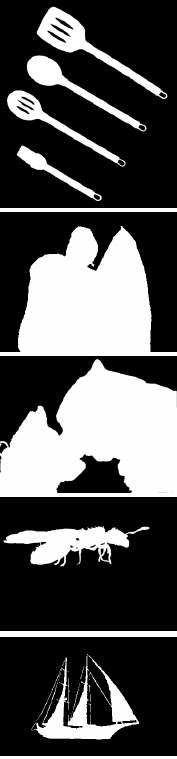}
		\centering\footnotesize{GT}
	\end{minipage}
	\begin{minipage}[t]{0.07\textwidth}
		\centering
		\includegraphics[scale=0.45]{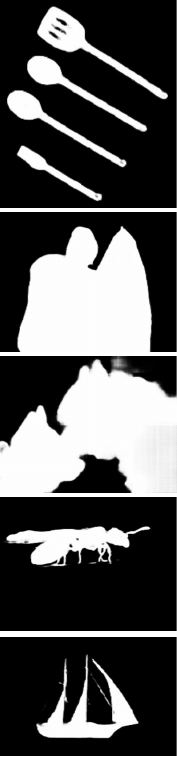}
		\centering\footnotesize{Fold with overlap}
	\end{minipage}
	\begin{minipage}[t]{0.07\textwidth}
		\centering
		\includegraphics[scale=0.45]{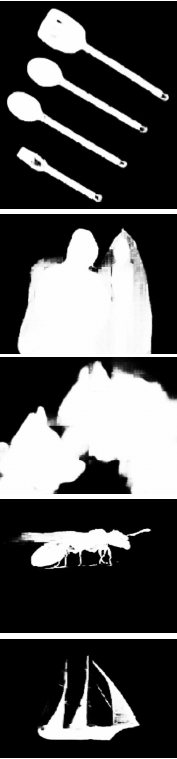}
		\centering\footnotesize{Fold}
	\end{minipage}
	\begin{minipage}[t]{0.07\textwidth}
		\centering
		\includegraphics[scale=0.45]{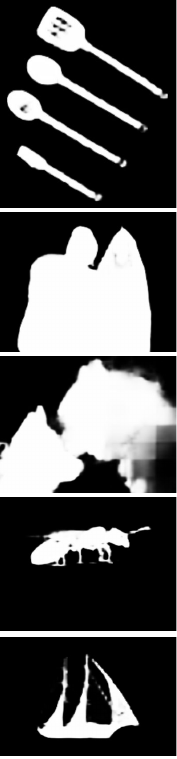}
		\centering\footnotesize{Bilinear}
	\end{minipage}
	%\begin{minipage}[t]{0.07\textwidth}
	%	\centering
	%	\includegraphics[scale=0.45]{figures/upsample_methods_convtrans.pdf}
	%	\centering\footnotesize{transpose conv}
	%\end{minipage}
	\begin{minipage}[t]{0.07\textwidth}
		\centering
		\includegraphics[scale=0.45]{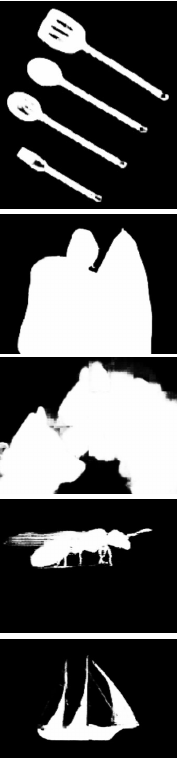}
		\centering\footnotesize{Pixel shuffle}
	\end{minipage}
	\begin{minipage}[t]{0.07\textwidth}\end{minipage}
	\par\;\caption{Visual comparisons of upsampling methods. 
		As can be observed, the prediction imaps obtained through upsampling with Fold and Pixel Shuffle exhibit undersaturation issues, while Bilinear interpolation results in a notable loss of local details. 
	}
	\label{fig:upsample}
\end{figure}
	
\begin{table}
	\centering
	\scriptsize
	\caption{Ablation study of different loss function settings with our M$^3$Net. }% $\mathcal{L}_{MS}$ denotes multilevel supervision. }
\label{tab:ablation_loss}
\setlength\tabcolsep{0mm}
\begin{tabular}{c|c|ccc|ccc|ccc}% 
	\hline
	\multirow{2}{*}{ID} & \multirow{2}{*}{Loss Settings} & \multicolumn{3}{c|}{DUTS} & \multicolumn{3}{c|}{ECSSD} & \multicolumn{3}{c}{HKU-IS}\\
	&& $M\downarrow$ & $S_m\uparrow$ & $F_\beta^w\uparrow$ & $M\downarrow$ & $S_m\uparrow$ & $F_\beta^w\uparrow$ & $M\downarrow$ & $S_m\uparrow$ & $F_\beta^w\uparrow$ \\
	\hline
	23 & $\mathcal{L}_{bce}$ & .039 & {.895} & {.835} & {.032} & {.93} & {.909} & {.029} & {.928} & {.903} \\
	5 & $\mathcal{L}_{bce} + \mathcal{L}_{iou}$ & \textbf{.036} & \textbf{.897} & \textbf{.849} & \textbf{.029} & \textbf{.931} & \textbf{.919} & \textbf{.026} & \textbf{.929} & \textbf{.913} \\
	%4& $\mathcal{L}_{api}$\cite{tracer} & \textbf{.036} & \textbf{.897} & {.847} & \textbf{.029} & {.927} & {.918} & {.027} & {.925} & {.908} \\
	24 & $\mathcal{L}_{wbce} + \mathcal{L}_{wiou}$\cite{F3Net} & .038 & {.895} & {.846} & {.03} & {.93} & {.917} & {.027} & {.927} & {.909} \\
	
	\hline    
\end{tabular}
\end{table}
Upsampling methods for features in the form of a sequence remain under studied. We compare several widely used upsampling methods with our M$^3$Net, including bilinear, pixel shuffle, and fold. The results are shown in Table \ref{tab:ablation_up} and Fig \ref{fig:upsample}. From the comparison result, we found that fold with overlap not only leads in evaluation metrics but also makes the predicted saliency map closest to the real one. 

\subsubsection{Evaluation of Loss Function}
We use different loss functions to train our M$^3$Net, and the results are shown in Table \ref{tab:ablation_loss}. We find that the model supervised by BCE loss is weak in $F_\beta^w$~\cite{6909433}, which may be caused by BCE ignoring the relations between pixels, while IoU loss based on the whole region, effectively made up for the deficiencies of BCE loss. $\mathcal{L}_{wBCE}$ and $\mathcal{L}_{wIoU}$~\cite{F3Net} assign different weights to different regions, aiming to let the model pay more attention to complex regions. However, as shown in Table \ref{tab:ablation_loss}, it has not brought visible improvement to the performance of our model. 
We speculate that this lack of improvement may stem from the model's pre-existing proficiency in attending to complex regions. 
Consequently, the emphasis on complex regions driven by weighted loss has the potential to impede the model's perception of non-complex regions. 
%The effectiveness of multilevel supervision was also confirmed (ID:6, 15), further demonstrating the rationality of our supervision strategy. 

%\subsection{Limitations}
%While our proposed model has achieved promising improvements in SOD tasks, it is important to acknowledge its limitations. Specifically, the mixed attention block incurs a relatively high computational cost, which limits its employment to smaller resolutions. Moreover, the performance of the model on other SOD tasks, such as those involving RGB-D and RGB-T, requires further investigation. We will carry on extensive research on these SOD tasks. 

\section{Conclusion}
In this paper, we propose a novel Transformer based network dubbed M$^3$Net for SOD. Considering the uniqueness and interdependence of multilevel features, we first propose the MIB to achieve the interaction between multilevel features and thus enhance the localized salient regions in low-level features. Secondly, we design the MAB which integrates the global and window self-attentions, aiming at modeling local context to refine the fine-grained details of the objects. Finally, we design a multistage decoder by employing the MIB and MAB blocks and optimize the multilevel features step by step. Our M$^3$Net model achieves state-of-the-art results on six challenging datasets without relying on heavy numerical computations, thus showing great potential for the SOD task in practical application.

%\section*{Acknowledgments}
%This should be a simple paragraph before the References to thank those individuals and institutions who have supported your work on this article.

\bibliographystyle{IEEEtran}
\bibliography{M3Net}

\vfill

\end{document}